\RequirePackage{amsmath}

\documentclass[a4paper]{article}

\usepackage{amssymb}
\usepackage{graphicx}

\usepackage{url}
\urldef{\mailsa}\path|{boettger, ulrich, steger}@mvtec.com|

\usepackage[pagebackref=true,breaklinks=true,colorlinks,bookmarks=false]{hyperref}

\usepackage[caption=false]{subfig}

\usepackage{pgfplots}
\usepackage{pgf,tikz}
\usepackage{mathrsfs}
\usetikzlibrary{calc}
\usetikzlibrary{arrows}

\tikzset{
  ctrlpoint/.style={%
    draw=gray,
    circle,
    inner sep=0,
    minimum width=1ex,
  }
}

\newenvironment{customlegend}[1][]{%
    \begingroup
    \csname pgfplots@init@cleared@structures\endcsname
    \pgfplotsset{#1}%
}{%
    \csname pgfplots@createlegend\endcsname
    \endgroup
}%

\def\addlegendimage{\csname pgfplots@addlegendimage\endcsname}

\begin{document}
\definecolor{myyellow}{rgb}{.9,.9,0.}
\definecolor{qqqqff}{rgb}{0.,0.,1.}
\definecolor{contourcolor}{rgb}{0.,0.,1.}
\definecolor{xdxdff}{rgb}{0.49019607843137253,0.49019607843137253,1.}

\title{Subpixel-Precise Tracking of Rigid Objects in Real-time}

%
%
\author{Tobias B\"ottger%
\and Markus Ulrich\and Carsten Steger \\
MVTec Software GmbH, Munich, Germany\\
\url{http://www.mvtec.com}}
%

{}

%
%
\maketitle

\begin{abstract}
We present a novel object tracking scheme that can track rigid objects in real time. The approach uses subpixel-precise image edges to track objects with high accuracy. It can determine the object position, scale, and rotation with subpixel-precision at around 80fps. The tracker returns a reliable score for each frame and is capable of self diagnosing
a tracking failure. Furthermore, the choice of the similarity measure makes the approach inherently robust against occlusion, clutter, and nonlinear illumination changes. 
We evaluate the method on sequences from rigid objects from the OTB-2015 and VOT2016 dataset and discuss its performance. The evaluation shows that the tracker is more accurate than state-of-the-art real-time trackers while being equally robust.
\end{abstract}

\section{Introduction}
Visual object tracking is a fundamental problem in computer vision that is concerned with estimating the 2D pose of an object in a video sequence. It has a wide range of applications, such as robotics, human-computer interaction, and visual surveillance \cite{danelljan_beyond_2016,henriques_high_speed_2015,lepetit_keypoint_2006}. 

To cover the large amount of applications, the performance of trackers is usually evaluated on very diverse, publicly available, benchmarks, such as  VOT2016 \cite{vot_2016}, OTB-2015 \cite{wu_otb_2015}, or MOT16 \cite{mot_2016}.
Although the videos in the benchmarks are very diverse, they do not necessarily cover specific applications, but rather try to be as general as possible. This leads to the fact that, in general, trackers are optimized towards their generalization capabilities and not to a specific application \cite{danelljan_beyond_2016,held_2016_learning,nam_2016_learning}. Furthermore, the objects in these datasets are manually labeled with either axis aligned or oriented bounding boxes and the accuracy of a tracker is measured by its bounding box overlap \cite{Kristan_2016_novel}. Hence, trackers with subpixel precise localization or ones not restricted to oriented or axis-aligned rectangles do not necessarily have higher overlap scores in the benchmarks. Nevertheless, many industrial applications such as autonomous driving or the visual monitoring of industrial production processes require a good localization accuracy in real-time. 
For example, when picking an object from a conveyor belt with a robot, the bounding box of the object is not sufficient.

\begin{figure}
\centering
\includegraphics[width=0.15\textwidth]{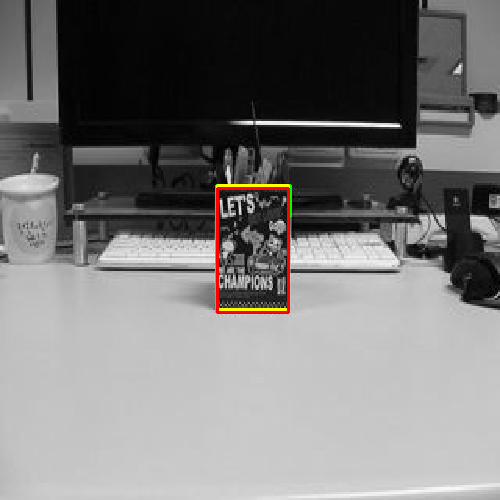}
\includegraphics[width=0.15\textwidth]{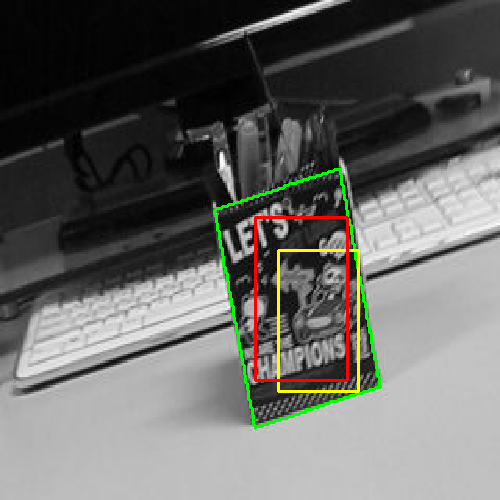}
\includegraphics[width=0.15\textwidth]{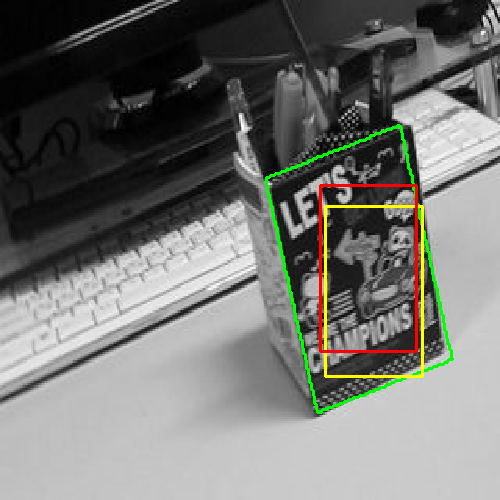}
\includegraphics[width=0.15\textwidth]{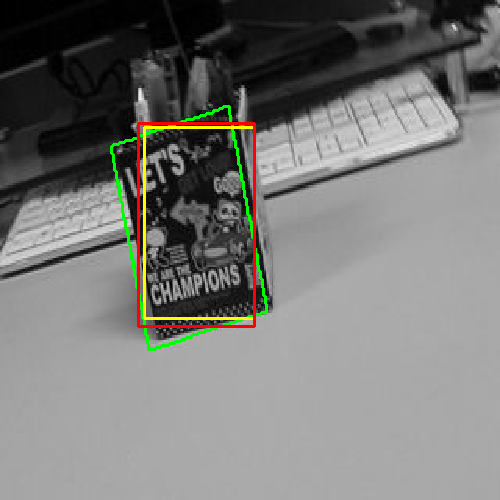}
\includegraphics[width=0.15\textwidth]{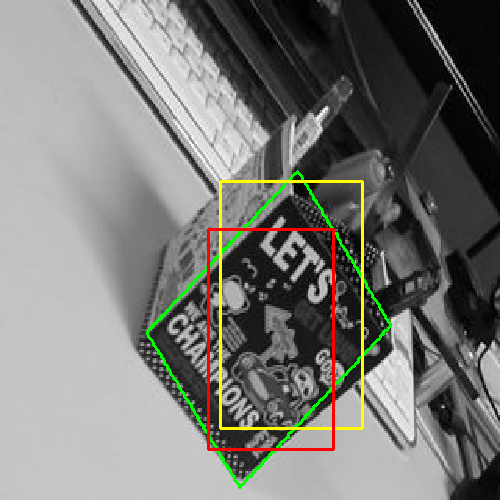}
\includegraphics[width=0.15\textwidth]{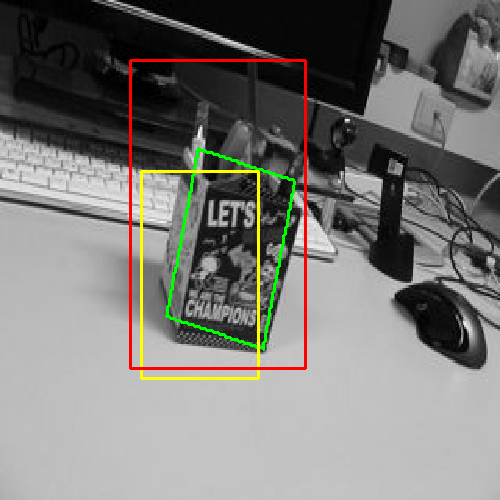}
\includegraphics[width=0.15\textwidth]{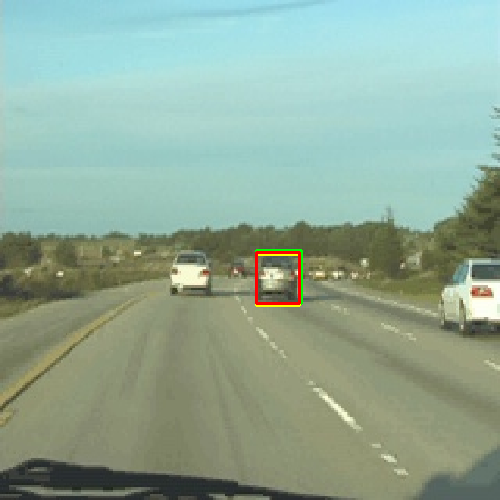}
\includegraphics[width=0.15\textwidth]{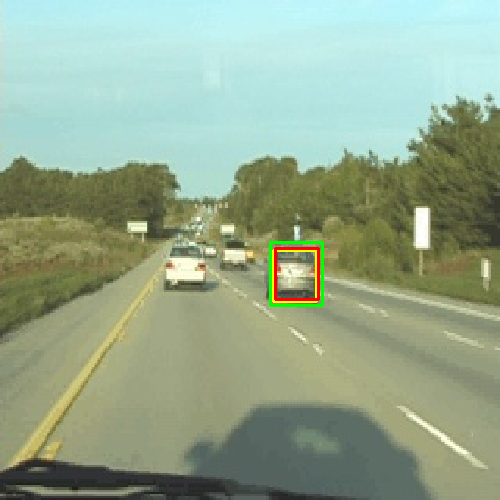}
\includegraphics[width=0.15\textwidth]{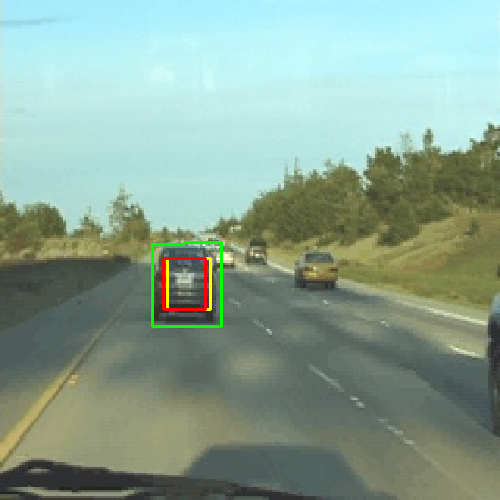}
\includegraphics[width=0.15\textwidth]{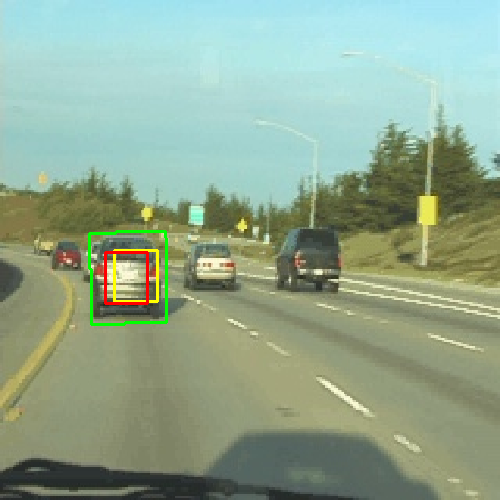}
\includegraphics[width=0.15\textwidth]{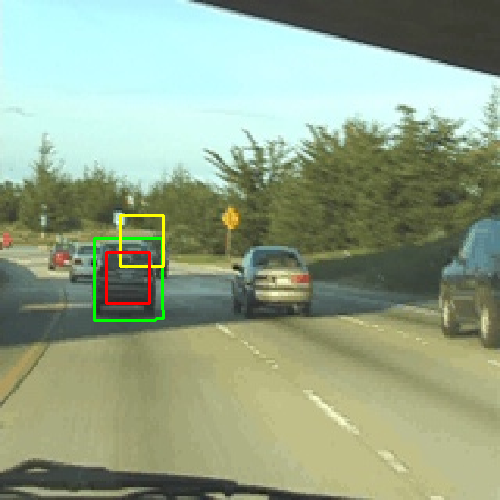}
\includegraphics[width=0.15\textwidth]{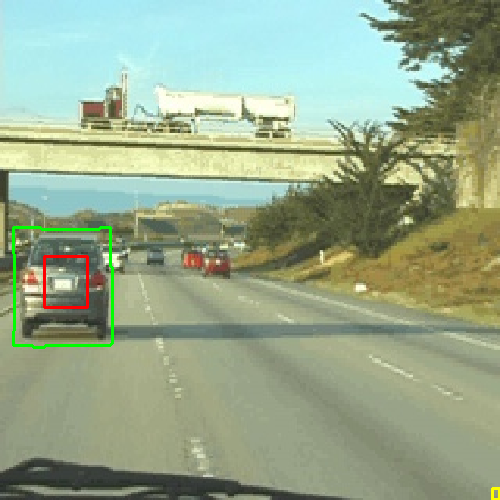}
\begin{tikzpicture}
\begin{customlegend}[legend entries={Correlation-based tracker \cite{danelljan_accurate_2014}\,\,,STAPLE \cite{bertinetto_2016_staple}\,\,,Our approach},
        legend columns=3]
    \addlegendimage{yellow,fill=black!100!yellow,sharp plot}
    \addlegendimage{red,fill=black!100!red,sharp plot}
    \addlegendimage{green,fill=black!100!green,sharp plot}
    \end{customlegend}
\end{tikzpicture}
\hfill
\caption{{\tt Vase} and {\tt Car24} from the OTB-2015 \cite{wu_otb_2015} benchmark. We compare our approach to two equally fast, state-of-the-art trackers: STAPLE \cite{bertinetto_2016_staple} and a scale adaptive correlation tracker \cite{danelljan_accurate_2014}. Our approach is able to accurately determine the position, scale, \emph{and} rotation of the objects, as opposed to just the axis aligned bounding box. The tracker is virtually drift-free in the second sequence (which has over 3000 frames). All three trackers run in real-time at 100 fps on both sequences.}
\label{fig:examplesequence}
\end{figure}

We present a real-time capable tracker that is able to determine the similarity transformation of a rigid object between frames. We leverage the fact that image edges can be determined with subpixel-precision to obtain a subpixel-precise object localization and do not restrict the object to a bounding box. The tracker returns a reliable score for each frame and is capable of self diagnosing a tracking failure.
The two examples sequences in \mbox{Fig. \ref{fig:examplesequence}.} show the localization quality of our approach and the fact that it is virtually drift-free for a rigid object in a sequence over 3000 frames. In the experiments section, we evaluate the method on further sequences from the OTB-2015 and VOT2016 dataset and comment on the performance characteristics of the presented approach. 

\section{Related Work}

Within visual object tracking, immense progress has been made in recent years. For example, the best performing tracker in the VOT-2014 challenge was only ranked $35$th in the 2016 challenge, with around half of the expected average overlap (AEO) of the VOT2016 winner \cite{vot_2016}. The great gain in performance is mostly due to the widespread adoption of discriminative learning methods such as discriminative correlation filters with complex features \cite{danelljan_beyond_2016,henriques_exploiting_2012,henriques_high_speed_2015,zhang_fast_2014} and deep convolutional neural networks (CNNs) for tracking \cite{held_2016_learning,nam_2016_learning,wang_2015_visual} (2015 VOT winner and the 2016 baseline for runner up). Although CNN-based trackers show impressive generalization properties and can cope with diverse sequences, most approaches are restricted to axis-aligned bounding boxes \cite{wang_2015_visual} and their accuracy is not on par with their robustness \cite{vot_2016}. Furthermore, the computational complexity of many approaches is infeasible and only few approaches are real-time capable with a high performance GPU \cite{held_2016_learning}, which is not an option in many industrial applications. 

In terms of speed, very robust trackers have emerged in the last few years which are real-time capable. Most of them build on discriminative correlation filters and extensions thereof \cite{bertinetto_2016_staple,danelljan_beyond_2016,henriques_exploiting_2012,henriques_high_speed_2015,zhang_fast_2014}. For example, 
the STAPLE tracker \cite{bertinetto_2016_staple} combines a HOG-based correlation filter with a model based on
color statistics and achieved the best real-time performance at the VOT-2016 challenge \cite{vot_2016}. Danelljan \emph{et al.} \cite{danelljan_beyond_2016} go beyond the ordinary  discriminative correlation framework and train
continuous convolution filters. Their approach performs on par with trackers based on CNNs and can be extended to subpixel-precise feature point tracking. Unfortunately, all of the mentioned approaches are restricted to axis aligned bounding boxes.

Similar to our approach, Lepetit and Fua \cite{lepetit_keypoint_2006} present a tracker that estimates the pose of rigid objects. Their keypoint-based recognition system is robust to occlusion and clutter, but requires an extensive offline training phase to generate the tracking model. In contrast to our approach, their tracking is restricted to textured objects that exhibit sufficient keypoints for reliable tracking.

As opposed to the above mentioned approaches, our approach is capable of estimating the position, scale, \emph{and} rotation of an arbitrarily shaped object in real-time and does not require an extensive offline training nor is it restricted to textured objects.


\section{Shape-Based Tracking}
Our tracking approach builds on the efficient shape-based object recognition technique of Steger \cite{steger_similarity_2001}. In the first frame, a shape-based model is generated from the arbitrarily shaped ROI of the detected or marked object. The model is used to determine the optimum object pose in the subsequent frames. After each successful tracking step, the model is updated and unstable points are filtered out and new points are added. The three steps (1) model generation, (2) model localization and, (3) model update are explained in more detail in the following section. Furthermore, we describe how the approach is made efficient and able to track most objects in the  VOT2016 \cite{vot_2016} and OTB-2015 \cite{wu_otb_2015} datasets at around 80fps without using the GPU on an IntelCore i7-4810 CPU @2.8GHz with 16GB of RAM with Windows 7 (x64). 

\subsection{Model Generation}
\begin{figure}
\centering
\subfloat[Example Image from the OTB-2015 benchmark \cite{wu_otb_2015}]
{
	\includegraphics[width=0.35\textwidth]{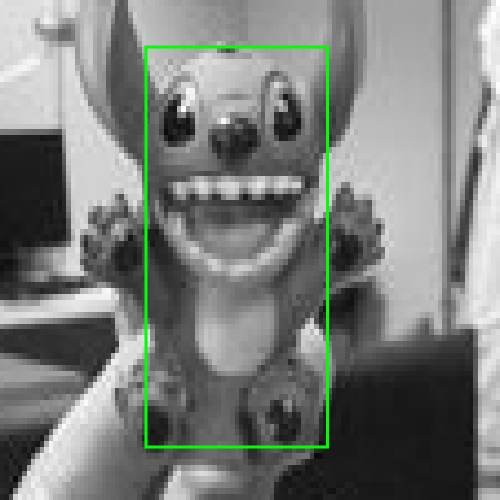}
}
\hfill
\subfloat[Model Points $p_i$ and their direction vectors]
{
  
  \begin{tikzpicture}[line cap=round,line join=round,>=latex,x=1.0cm,y=1.0cm, scale=0.52, every node/.style={scale=0.7}]
  \clip(-2,-1) rectangle (6,7);
  \node (p1) [ctrlpoint,label=60:$p_1$] at (4,4) {};
  \node (p2) [ctrlpoint,label=90:$p_2$] at (3.5,5.3) {};
  \node (p3) [ctrlpoint,label=45:$p_3$] at (2,6) {};
  \node (p4) [ctrlpoint,label=90:$p_4$] at (0.62,5.4) {};
  \node (p5) [ctrlpoint,label=135:$p_5$] at (0.05,4.) {};
  \node (p6) [ctrlpoint,label=180:$p_6$] at (0.61,2.6) {};
  \node (p7) [ctrlpoint,label=45:$p_7$] at (3.0,1.56) {};
  \node (p8) [ctrlpoint,label=45:$p_8$] at (1.0,1.50) {};
  \node (p9) [ctrlpoint,label=90:$p_9$] at (2,0) {};
  \node (p10) [ctrlpoint,label=45:$p_{10}$] at (2.0,2.0) {};
  \node (p11) [ctrlpoint,label=45:$p_{11}$] at (0.8,3.4) {};
  \node (p12) [ctrlpoint,label=45:$p_{12}$] at (3.0,3.4) {};
  \node (p13) [ctrlpoint,label=0:$p_{13}$] at (3.4,2.64) {};
  \coordinate (h1) at (1.13,2.25);
  \coordinate (h2) at (2.78,2.25);
  \coordinate (h3) at (1.3,0.5);
  \coordinate (h4) at (2.82,0.5);
  \coordinate (e1) at (1.1,5.25);
  \coordinate (e2) at (1.4,4.6);
  \coordinate (e3) at (0.8,4.5);
  \coordinate (e4) at (2.85,5.3);
  \coordinate (e5) at (3.3,4.65);
  \coordinate (e6) at (2.78,4.6);
  \coordinate (n1) at (2,4.75);
  \coordinate (n2) at (1.6,4.2);
  \coordinate (n3) at (2.5,4.3);
  \coordinate (n4) at (2.0,3.7);
  \coordinate (m1) at (0.6,3.4);
  \coordinate (m2) at (3.2,3.4);
  \coordinate (m3) at (2.9,2.7);
  \coordinate (m4) at (1.1,2.7);
  \draw [->] (p1) -- (4.62,4.08);
  \draw [->] (p2) -- (3.98,5.74);
  \draw [->] (p3) -- (1.98,6.6);
  \draw [->] (p4) -- (0.2,5.8);
  \draw [->] (p5) -- (-0.70,3.93);
  \draw [->] (p6) -- (0.2,2.2);
  \draw [->] (p7) -- (3.6,1.59);
  \draw [->] (p8) -- (0.3,1.4);
  \draw [->] (p9) -- (2,-0.5);
  \draw [->] (p10) -- (2.0,1.3);
  \draw [->] (p11) -- (0.8,4.0);
  \draw [->] (p12) -- (3.0,4.0);
  \draw [->] (p13) -- (3.8,2.);
  \draw [color=contourcolor, xshift=0cm] plot [smooth cycle] coordinates {(p1) (p2) (p3) (p4) (p5) (p6) (p10) (p13) (p1)};
  \draw [color=contourcolor, xshift=0cm] plot [smooth cycle] coordinates {(m1) (p11) (p12) (m2) (m3) (m4)};
  \draw [color=contourcolor, xshift=0cm] plot [smooth] coordinates {(h1) (p8) (h3) (p9) (h4) (p7) (h2)};
  \draw [color=contourcolor, xshift=0cm] plot [smooth cycle] coordinates {(e1) (e2) (e3)};
  \draw [color=contourcolor, xshift=0cm] plot [smooth cycle] coordinates {(e4) (e5) (e6)};
  \draw [color=contourcolor, xshift=0cm] plot [smooth cycle] coordinates {(n1) (n3) (n4) (n2)};
  \end{tikzpicture}
}
\hfill
\caption{The model in (b) is generated from the ROI in the first input frame displayed in (a). The model point are generated from non-maximum suppressed edges with a high enough edge amplitude (see \cite{ulrich2011system}). The model in (b) is simplified for better visualization. The real model has over 400 points.}
\label{fig:modelexample}
\end{figure}
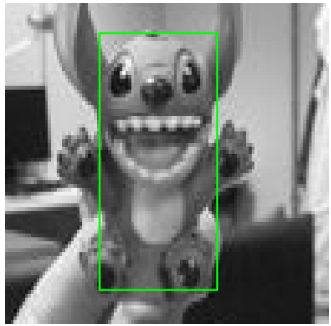
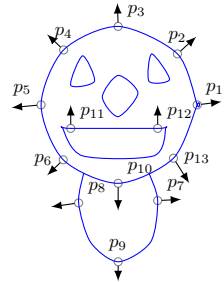

In the first frame, the tracking model $\mathcal{M}$ is constructed from the ROI of the automatically detected or manually marked object. The model consists of a set of $n$ points $p_i = (x_i,y_i)^T$ and their corresponding direction vectors $d_i=(t_i,u_i)^T$:
\begin{equation} \label{eq:model}
\mathcal{M} = \{(p_i,d_i) \in\mathbb{R}^2\times\mathbb{R}^2,\text{for } i  = 1,\dots, n\}.
\end{equation}
In a first step, point candidates are extracted by applying a threshold on the Sobel filter edge amplitude of the input ROI. To thin out the number of points, non-maximum suppression is applied with automatically estimated thresholds, see the patent \cite{ulrich2011system} for details. The remaining points can then be refined to subpixel-precision which is described in more detail in Chapter~3.3 of \cite{steger_unbiased_1998}. The coordinates of the model points are all expressed relative to an arbitrary reference point. We use the center of the bounding box for simplicity. An exemplary model is displayed in \mbox{Fig. \ref{fig:modelexample}}.

Please note, that we do not have a training phase of our model. The model consists of a single set of points and their directions. The model transformation to different poses is done on the fly in the localization step. 

\subsection{Model Localization}
The model localization essentially amounts to finding the best matching candidate within the target image in a template matching framework. Hence, we compare a transformed model to the target image at a particular location by a similarity measure. By setting a minimal required similarity, we are able to avoid a very large number of computations.

In a first step, we calculate a direction vector for each pixel within the current frame and identify them as $e_{x,y} = (v_{x,y},w_{x,y})$. We can then evaluate the similarity of the tracking model $\mathcal{M}$ to the current frame at various image locations and for different transformations of the tracking model.
The location and transformation parameters with the highest similarities are the most probable object locations. 
The similarity transformation of a model point is given by:
\begin{equation} \label{eq:transformation}
p_i' =  \underbrace{\begin{pmatrix}\sigma \cos\theta & -\sigma\sin\theta \\
\sigma \sin\theta & \,\,\, \sigma \cos\theta \end{pmatrix}}_{T_{\theta,\sigma}} p_i + \begin{pmatrix} x_t \\ y_t \end{pmatrix},
\end{equation}
where $\theta$ and $\sigma$ are the rotation and scale parameters, respectively.
Similarly, the transformed direction vectors are obtained by
\begin{equation} \label{eq:transformation_direction}
d_i' = (T_{\theta,\sigma}^{-1})^T d_i.
\end{equation}

As similarity measure we use the normalized sum of dot products of the normalized direction vectors of the transformed model and the target image:

\begin{equation} \label{eq:similarity}
s(x_t,y_t,\theta, \sigma)_\mathcal{M} = \frac{1}{n} \left\vert \sum_{i=1}^{n} \frac{\left\langle d_i',e_{p_i'} \right\rangle}{\Vert d_i' \Vert \cdot \Vert e_{p_i'}\Vert}\right\vert,
\end{equation}
with $s :\mathbb{R}^4 \rightarrow [0,1]$. The similarity measure is robust to occlusion, clutter, non-linear illumination changes, and a moderate amount of defocusing \cite{steger_similarity_2001}. The robustness to non-linear illumination change comes from the fact that all direction vectors are scaled to unit-length. The robustness to occlusion comes from the fact that missing points in the target image will, on average, contribute nothing to the sum of \eqref{eq:similarity}. Similarly, clutter lines or points in the target image do not only need to coincide with the sparse set of model points, but also need to have a similar direction vectors to contribute to the similarity. 

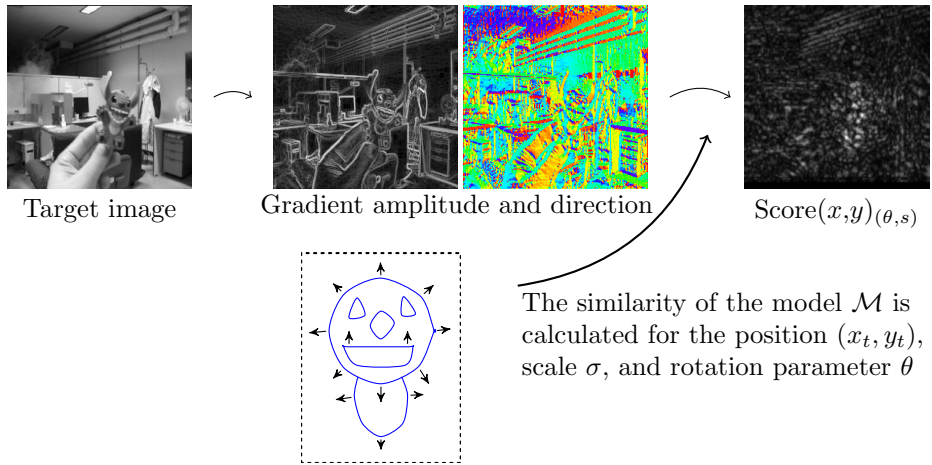
\begin{figure}
\centering
\begin{tikzpicture}
[
	 box/.style={rectangle,draw=black,thick, minimum size=1cm},
]

\node[inner sep=0pt] (target) [label=270:Target image] at (-3,0)
{\includegraphics[width=0.2\textwidth]{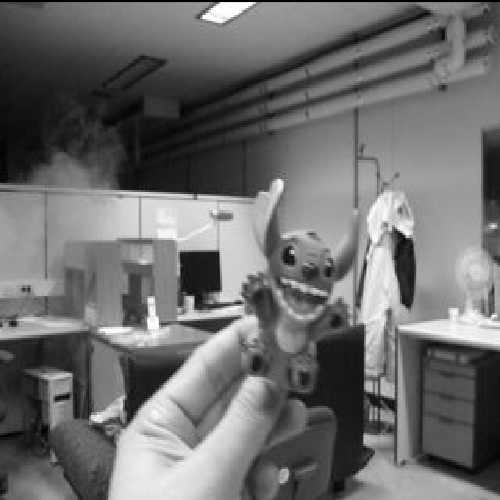}};

\path[->] (-1.5,0) edge[bend left] node[above] {} (-1,0);   

\begin{scope}
	\node[inner sep=0pt] (magnitude) [] at (0.5,0)
	{\includegraphics[width=0.2\textwidth]{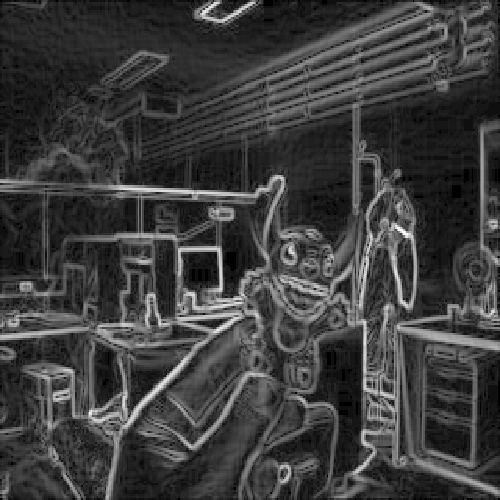}};
	\node[inner sep=0pt] (direction) [] at (3,0)
	{\includegraphics[width=0.2\textwidth]{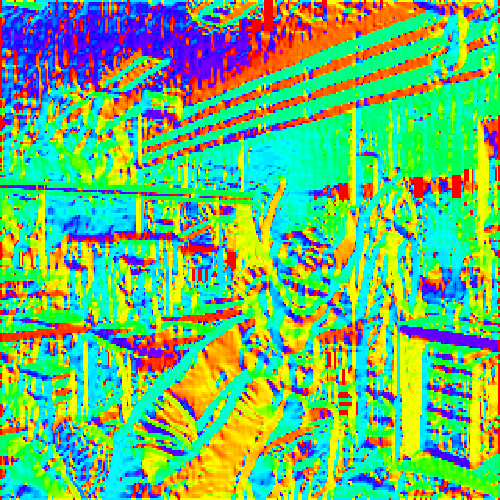}};
\end{scope}

\path[->] (4.5,0) edge[bend left] node[above] {} (5.3,0); 

\node[inner sep=0pt] (score) [label=270:$\text{Score}(x{,}y)_{(\theta,s)}$] at (6.7,0)
{
\includegraphics[width=0.2\textwidth]{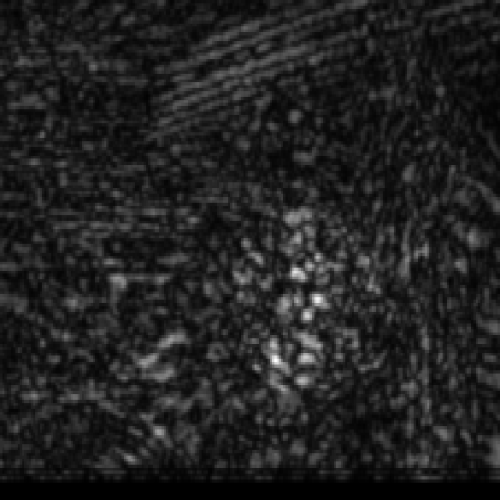}
};

\path[->,thick, bend right] (2.5,-2.5) edge[left] (5.0,-0.5); 

\node [label={[align=left]The similarity of the model $\mathcal{M}$ is \\calculated for the position $(x_t,y_t)$, \\scale $\sigma$, and rotation parameter $\theta$}] at (5.2,-4) {};

\node [label={[align=left]Gradient amplitude and direction}] at (1.7,-1.85) {};

\mbox{
\begin{scope}
[line cap=round,line join=round,>=stealth',yshift=-4.5cm, scale=0.35, every node/.style={scale=0.35}]
	\clip(-1,-1) rectangle (5,7);
	\node (p1) [] at (4,4) {};
	\node (p2) [] at (3.5,5.3) {};
	\node (p3) [] at (2,6) {};
	\node (p4) [] at (0.62,5.4) {};
	\node (p5) [] at (0.05,4.) {};
	\node (p6) [] at (0.61,2.6) {};
	\node (p7) [] at (3.0,1.56) {};
	\node (p8) [] at (1.0,1.50) {};
	\node (p9) [] at (2,0) {};
	\node (p10) [] at (2.0,2.0) {};
	\node (p11) [] at (0.8,3.4) {};
	\node (p12) [] at (3.0,3.4) {};
	\node (p13) [] at (3.4,2.64) {};
	\coordinate (h1) at (1.13,2.25);
	\coordinate (h2) at (2.78,2.25);
	\coordinate (h3) at (1.3,0.5);
	\coordinate (h4) at (2.82,0.5);
	\coordinate (e1) at (1.1,5.25);
	\coordinate (e2) at (1.4,4.6);
	\coordinate (e3) at (0.8,4.5);
	\coordinate (e4) at (2.85,5.3);
	\coordinate (e5) at (3.3,4.65);
	\coordinate (e6) at (2.78,4.6);
	\coordinate (n1) at (2,4.75);
	\coordinate (n2) at (1.6,4.2);
	\coordinate (n3) at (2.5,4.3);
	\coordinate (n4) at (2.0,3.7);
	\coordinate (m1) at (0.6,3.4);
	\coordinate (m2) at (3.2,3.4);
	\coordinate (m3) at (2.9,2.7);
	\coordinate (m4) at (1.1,2.7);
	\draw [->] (p1) -- (4.62,4.08);
	\draw [->] (p2) -- (3.98,5.74);
	\draw [->] (p3) -- (1.98,6.6);
	\draw [->] (p4) -- (0.2,5.8);
	\draw [->] (p5) -- (-0.70,3.93);
	\draw [->] (p6) -- (0.2,2.2);
	\draw [->] (p7) -- (3.6,1.59);
	\draw [->] (p8) -- (0.3,1.4);
	\draw [->] (p9) -- (2,-0.5);
	\draw [->] (p10) -- (2.0,1.3);
	\draw [->] (p11) -- (0.8,4.0);
	\draw [->] (p12) -- (3.0,4.0);
	\draw [->] (p13) -- (3.8,2.);
	\draw [color=contourcolor, xshift=0cm] plot [smooth cycle] coordinates {(p1) (p2) (p3) (p4) (p5) (p6) (p10) (p13) (p1)};
	\draw [color=contourcolor, xshift=0cm] plot [smooth cycle] coordinates {(m1) (p11) (p12) (m2) (m3) (m4)};
	\draw [color=contourcolor, xshift=0cm] plot [smooth] coordinates {(h1) (p8) (h3) (p9) (h4) (p7) (h2)};
	\draw [color=contourcolor, xshift=0cm] plot [smooth cycle] coordinates {(e1) (e2) (e3)};
	\draw [color=contourcolor, xshift=0cm] plot [smooth cycle] coordinates {(e4) (e5) (e6)};
	\draw [color=contourcolor, xshift=0cm] plot [smooth cycle] coordinates {(n1) (n3) (n4) (n2)};
	
	\draw[thick,dotted]     ($(-1,-1)$) rectangle ($(5.0,6.95)$);
\end{scope}
}

\end{tikzpicture}
\caption{In a first step, the gradient amplitudes and direction of the target image are calculated. The amplitudes are required for the subpixel-precise refinement of the object position. The maximum similarity of the model $\mathcal{M}$ from Fig. \ref{fig:modelexample}. is calculated for the position, scale and angle within the discretized 4d search space.}
\label{fig:flowchart}
\end{figure}

The localization process is visualized in the flowchart of \mbox{Fig. \ref{fig:flowchart}}. At this point the optimal position, angle, and scale are determined with pixel accuracy. In the following subsection, we display how the 4d optima $(\tilde x_t,\tilde y_t,\tilde \theta, \tilde \sigma)$ is refined to subpixel accuracy. 

\subsection{Subpixel-precise Refinement}
The accuracy of the localization step depends on the chosen discretization of $\sigma$ and $\theta$ as well as the pixel resolution. To refine the match, we apply the concept of the 4d facet model. We approximate the 4d parameter space by calculating a second order Taylor polynomial around the $3\times3\times3\times3$ best match and extracting the maximum of this polynomial \cite{steger_similarity_2001}. 

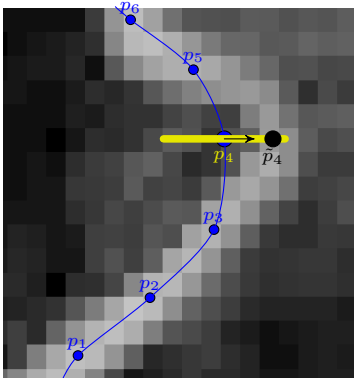
\begin{figure}
\centering

\begin{tikzpicture}[line cap=round,line join=round,>=stealth',x=1.0cm,y=1.0cm, scale=0.40]
\clip(0.5,0.5) rectangle (12,13);

\node[anchor=south west,inner sep=0] at (0,0) {\includegraphics[width=0.42\textwidth]{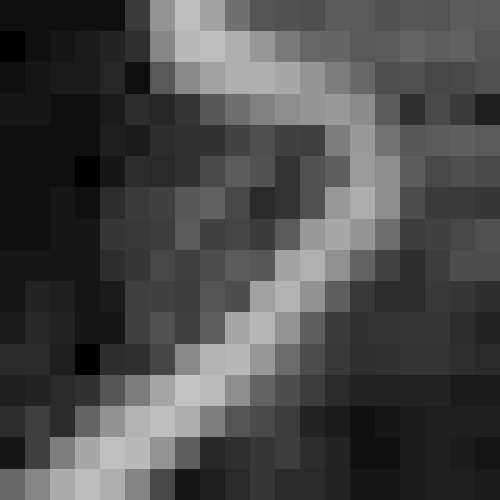}};

\coordinate (h1) at (2.2486056594196715,0);
\coordinate (p1) at (2.9486056594196715,1.2388217387241707);
\coordinate (p2) at (5.3171883947384275,3.156245857791734);
\coordinate (p3) at (7.422595270577323,5.41203893904769);
\coordinate (p4) at (7.7609642327657165,8.419763047388965);
\coordinate (p5) at (6.745857346200535,10.713152679999189);
\coordinate (p6) at (4.678047021715907,12.36740093958689);
\coordinate (h2) at (4.178047021715907,12.8);

\coordinate (b4) at (9.3609642327657165,8.419763047388965);

\coordinate (v4) at (8.7609642327657165,8.4197630473889);
\coordinate (h3) at (5.7609642327657165,8.4197630473889);
\coordinate (h4) at (9.7609642327657165,8.4197630473889);

\begin{scriptsize}
\draw [fill=qqqqff] (p1) circle (4.5pt);
\draw[color=qqqqff] (p1)+(0.0,0.4) node {$p_1$};
\draw [fill=qqqqff] (p2) circle (4.5pt);
\draw[color=qqqqff] (p2)+(0.0,0.4) node {$p_2$};
\draw [fill=qqqqff] (p3) circle (4.5pt);
\draw[color=qqqqff] (p3)+(0.0,0.4) node {$p_3$};
\draw [fill=qqqqff] (p4) circle (7.5pt);
\draw[color=myyellow] (p4)+(0.0,-0.6) node {$p_4$};
\draw [fill=qqqqff] (p5) circle (4.5pt);
\draw[color=qqqqff] (p5)+(0.0,0.4) node {$p_5$};
\draw [fill=qqqqff] (p6) circle (4.5pt);
\draw[color=qqqqff] (p6)+(0.0,0.4) node {$p_6$};

\draw [color=qqqqff] plot [smooth,line width=0.5mm] coordinates {(h1) (p1) (p2) (p3) (p4) (p5) (p6) (h2)};

\draw [color=myyellow,line width=1mm] plot [smooth] coordinates {(h3) (p4) (h4)};

\draw [->] (p4) -- (v4);

\draw [postaction={fill=black}] (b4) circle (7.5pt);
\draw[color=black] (b4)+(0.0,-0.6) node {$\tilde p_4$};

\end{scriptsize}
\end{tikzpicture}

\hfill
\caption{The model is adapted to transformations that cannot be captured by a similarity transformation of the complete model. After the optimal model position, scale, and rotation have been determined, each point searches for its best match along a 1d search line perpendicular to its image tangent. The length of the search line is variable, but has a significant impact on the runtime.}
\label{fig:modelupdate}
\end{figure}

To further improve the localization accuracy and the robustness of the tracking to small model deformations and transformations that cannot be captured by a similarity transformation, we use a modified version of the least squares refinement described in \cite{ulrich_2002_performance}. The least-squares refinement assumes a good initial approximation of the current transformation and improves the global similarity transformation for all points. For each model point $p_i$, the best point match in the direction of $\pm d'_i$ is determined. The concept is displayed in \mbox{Fig. \ref{fig:modelupdate}} and explained in more detail in \cite{ulrich_2002_performance}. In contrast to \cite{ulrich_2002_performance}, we do not restrict the length of the search line to 1, but rather allow an arbitrarily long search line. We found that a single least squares iteration was sufficient for most tracking sequences.
\subsection{Model Update}
After we have successfully refined the object pose to subpixel-precision, we conduct one final search for the best corresponding target image point for each model point in the direction of $\pm d'_i$. This time we do not update the global similarity transformation of the model, but rather update the relative positions of the points themselves. This improves how well the model will fit to the target image at future timesteps.

In the example shown in \mbox{Fig. \ref{fig:modelupdate}}, we shift $p_4$ towards the best match $\tilde p_4$ that is found along the yellow line. We regularize the model update with a parameter $\lambda$ to be more robust to noisy object deformations. At frame $t$ we update each point $p_i^t$ with its best match $\tilde p_i^t$ such that:
\begin{equation}
p_4^{t+1} = p_4^{t} + \lambda \tilde p_4^{t}.
\end{equation}

The update step does not only allow our approach to capture small model deformations, but also weakens the restriction of our approach to similarity transformations of the model, consequently projective transformations that increment over time may be captured by locally deforming the model points. 

Please note that in tracking settings the model update step always needs to find the balance between keeping the localization accuracy high and generalizing well to new representations of the model. In our approach, too large parameter values of $\lambda$ may add drift and can lead to a degeneration of the tracking model if no extra care is taken. Nevertheless, since the model transformation is determined with subpixel-precision, even long sequences with over 3000 frames, like the one in the example displayed in \mbox{Fig. \ref{fig:examplesequence}.}, do not drift significantly.

\begin{figure}
\centering
\includegraphics[width=0.15\textwidth]{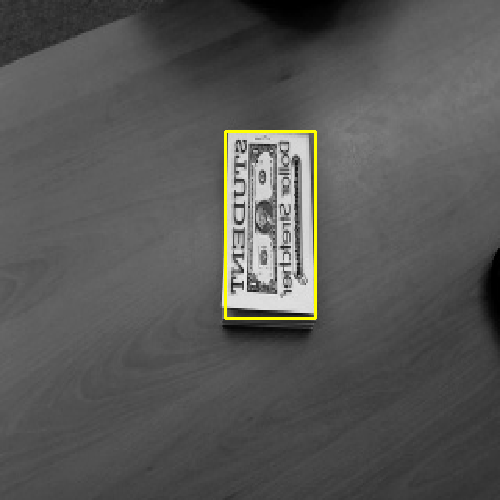}
\includegraphics[width=0.15\textwidth]{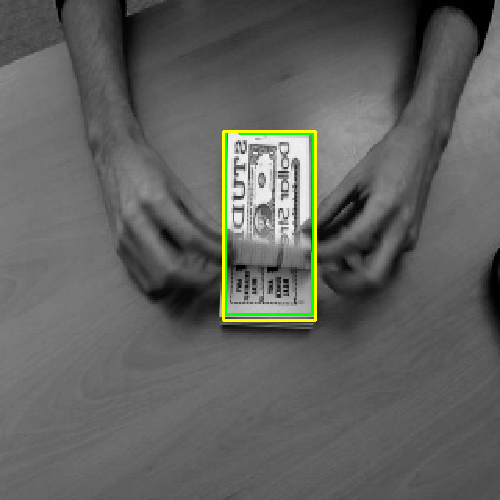}
\includegraphics[width=0.15\textwidth]{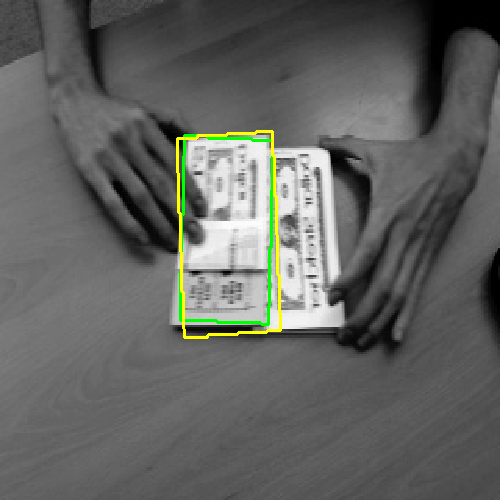}
\includegraphics[width=0.15\textwidth]{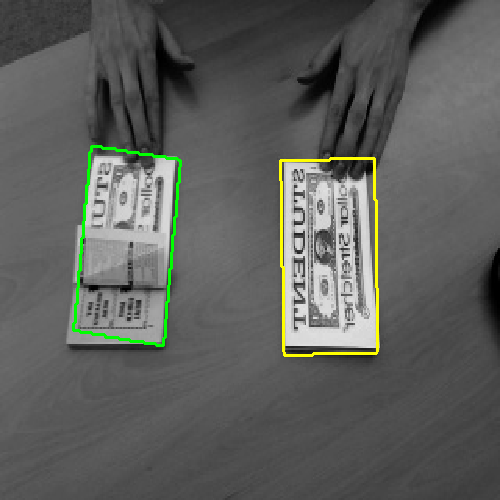}
\includegraphics[width=0.15\textwidth]{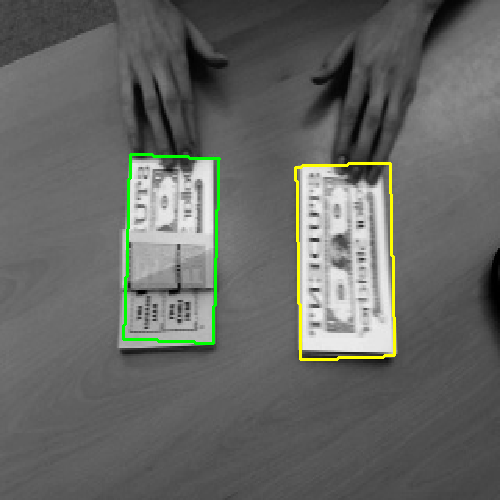}
\includegraphics[width=0.15\textwidth]{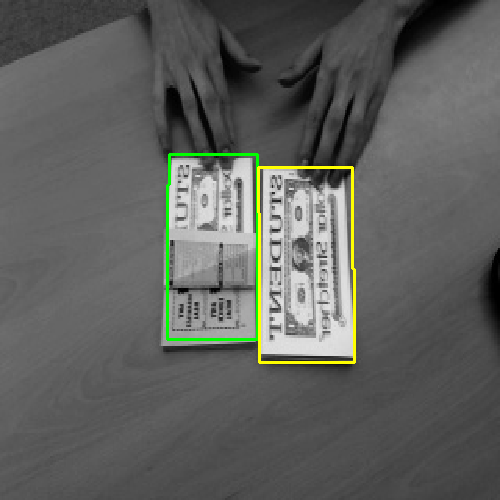}
\begin{tikzpicture}
\begin{customlegend}[legend entries={$\lambda=0.0$\,\,,$\lambda=0.7$},
        legend columns=2]
    \addlegendimage{yellow,fill=black!100!yellow,sharp plot}
    \addlegendimage{green,fill=black!100!green,sharp plot}
    \end{customlegend}
\end{tikzpicture}
\hfill
\caption{Sequence {\tt Coupon} from OTB-2015. If no model update is performed ($\lambda=0.0$) the tracker jumps to the wrong dollar note when the folded top note is moved. If the model update parameter $\lambda$ is set high enough, the model is updated when the note is folded and the tracker succeeds.}
\label{fig:exampledeformation}
\end{figure}

During the tracking process, we further monitor how often every model point is found. This step enables us to identify points that are not significantly contributing to the model localization and to remove them. These points may have either emerged from poorly initialized points in the first frame or by parts of the object becoming occluded or changing in later frames. To prevent us from deleting all points and degenerating the model, we sample new points in sparse areas of the model after a successful tracking step. This allows us to capture newly emerging object edges. The example in \mbox{Fig. \ref{fig:exampledeformation}.} shows how the model update helps to capture deformations of the object.

\subsection{Implementation Details}
In tracking we do not need to search for the model in the target image exhaustively in each frame. To reduce the workload, we restrict the possible parameter values of $(\theta,\sigma) \in [\theta_c-0.1,\theta_c+0.1] \times [\sigma_c-0.2,\sigma_c+0.2]$, where $\theta_c$ and $\sigma_c$ refer to the current object rotation and scale. Furthermore, for the translation, we define a circular search region with a radius of $1/2$ of the object diagonal. Although the search space was adequate for all of the test sequences, the size of the search region may be increased for fast moving objects.

If no parameter set of $(x_t,y_t,\theta, \sigma)$ that has a score $>s_{\text{min}}$ is found in a frame, we increase the search region and the parameter ranges of $(\theta, \sigma)$ for the next frames successively. As soon as we find the object again, we reset the search parameters to their initial value. To prevent the workload from becoming too large, we decrease the discretization of the search space when it becomes bigger. Although this decreases the accuracy, it gives us the chance to re-detect lost objects and improve the accuracy in the subsequent frames.

To achieve further speed-ups, we stop calculating a score for a model transformation as soon as it cannot reach a predefined minimal score $s_{\text{min}}$ anymore. To obtain an even larger speed-up it is possible to be even stricter, please refer to \cite{ulrich_2002_performance} for further details.
\begin{figure}
\centering
\includegraphics[width=0.15\textwidth]{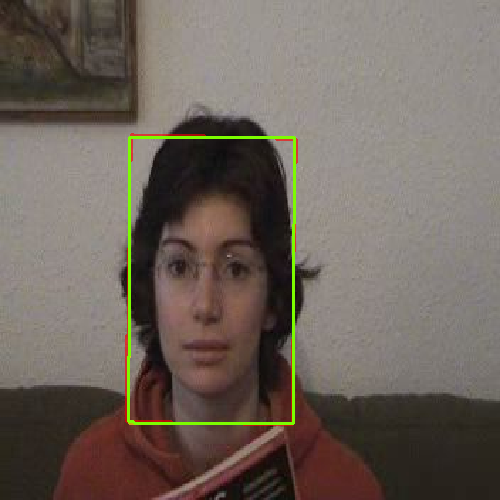}
\includegraphics[width=0.15\textwidth]{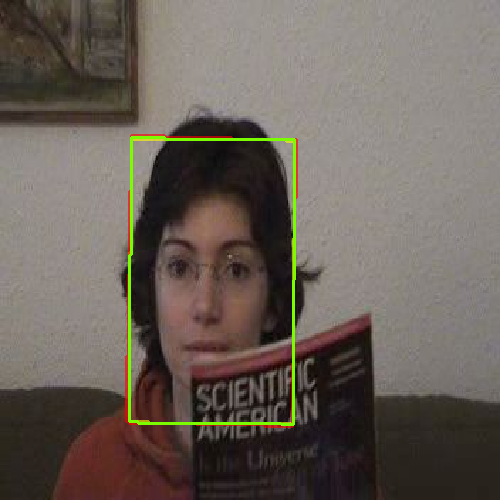}
\includegraphics[width=0.15\textwidth]{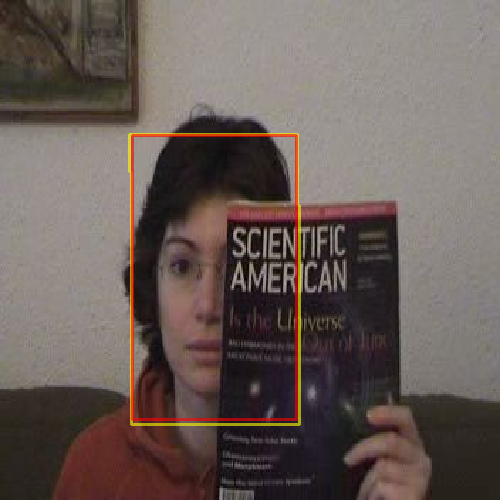}
\includegraphics[width=0.15\textwidth]{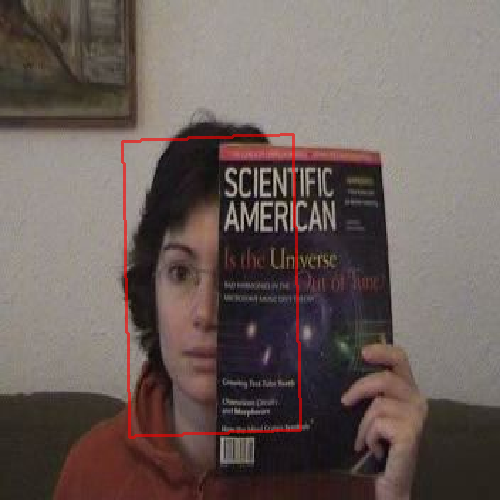}
\includegraphics[width=0.15\textwidth]{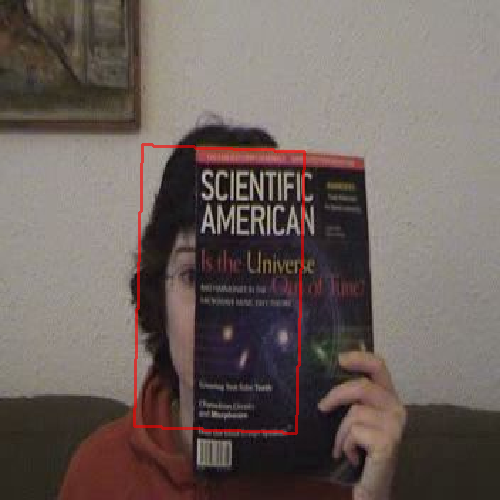}
\includegraphics[width=0.15\textwidth]{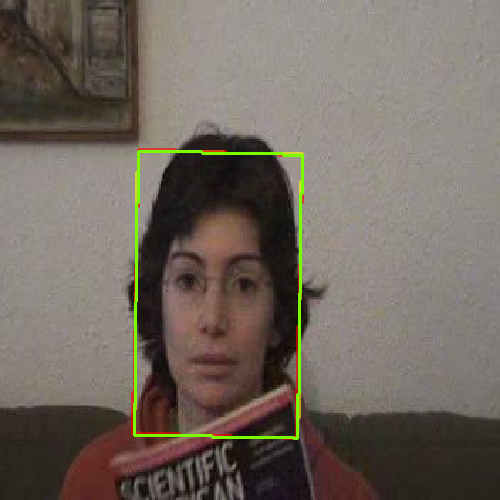}
\begin{tikzpicture}
\begin{customlegend}[legend entries={$s_{\text{min}}=0.4$\,\,,$s_{\text{min}}=0.6$\,\,,$s_{\text{min}}=0.8$},
        legend columns=3]
    \addlegendimage{red,fill=black!100!red,sharp plot}
    \addlegendimage{yellow,fill=black!100!yellow,sharp plot}
    \addlegendimage{green,fill=black!100!green,sharp plot}
    \end{customlegend}
\end{tikzpicture}
\hfill
\caption{The results for 3 different parameters of $s_{\text{min}}$ are displayed for the {\tt FaceOcc1} sequence from OTB-2015. The value of $s_{\text{min}}$ is an indicator of how much the object is allowed to be occluded. Lower values improve the robustness to occlusion but also require more time, since more score values need to be computed. For $s_{\text{min}}=0.8$, the object is not detected for the third to fifth image, while for $s_{\text{min}}=0.6$ it is lost for the fourth and fifth image. However, all approaches recover when the occlusion ends.}
\label{fig:exampleminscore}
\end{figure}

The parameter $s_{\text{min}}$ gives a good estimation of the allowed object occlusion. If half of the model points are occluded in the target image, the maximum score that can be obtained is $0.5$. This is displayed in \mbox{Fig. \ref{fig:exampleminscore}.}, where the face is essentially occluded by over $50\%$ and hence, only values of $s_{\text{min}}<0.5$ are able to track the object through all frames. Please note that a low value of $s_{\text{min}}$ increases the number of points for which the score needs to be calculated and hence has a negative impact on the runtime.

Further speed-ups can be obtained by only using a fixed number of points for tracking the object. Before each localization step, a random subset of points is selected from the model $\mathcal{M}$ and used for tracking. Although some accuracy may be lost, the execution time can be reduced.  

\section{Experiments}

The restriction to rigid objects and subpixel-precise localization makes it difficult to compare the approach to the vast majority of existing schemes in general. First of all, the data of existing benchmarks, such as VOT2016 \cite{vot_2016} and OTB-2015 \cite{wu_otb_2015}, are not annotated with sufficient accuracy and focus on robust, generalizable tracking. Hence, we focus our evaluation on selected sequences of rigid objects from both datasets and point out the strengths and weaknesses of the proposed approach.

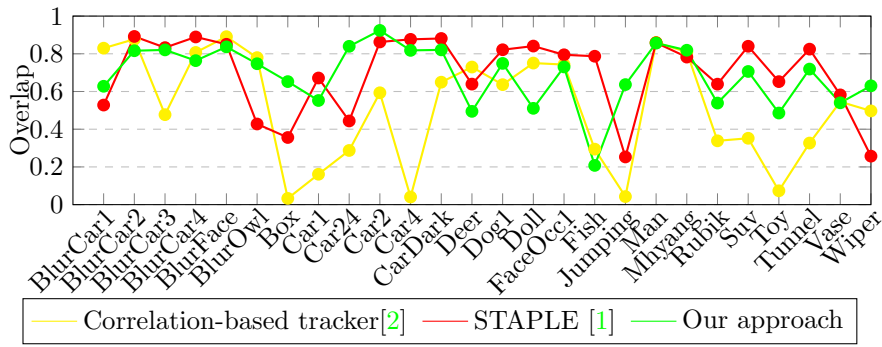
\begin{figure}
\centering
\begin{tikzpicture}
\begin{axis}[
	scale only axis=true,
	width=10.5cm, height=2.5cm,
	x label style={at={(axis description cs:0.5,0.1)},anchor=north},
	y label style={at={(axis description cs:0.08,.5)},anchor=south},
    ylabel={Overlap},
    xmin=0, xmax=26,
    ymin=0, ymax=1.00,
    xtick={1,2,3,4,5,6,7,8,9,10,11,12,13,14,15,16,17,18,19,20,21,22,23,24,25,26},
    xticklabels={BlurCar1,BlurCar2,BlurCar3,BlurCar4,BlurFace,BlurOwl,Box,Car1,Car24,Car2,Car4,CarDark,Deer,Dog1,Doll,FaceOcc1,Fish,Jumping,Man,Mhyang,Rubik,Suv,Toy,Tunnel,Vase,Wiper},
    ytick={0,.20,.40,.60,.80,1.00},
    x tick label style={rotate=45, anchor=north east, inner sep=0mm},
    legend pos=north west,
    ymajorgrids=true,
    grid style=dashed,
]
 
\addplot [color=yellow,thick,mark=*,mark options=solid] table {avg_overlap_kcf.txt};
\addplot [color=red,thick,mark=*,mark options=solid] table {avg_overlap_staple.txt};
\addplot [color=green,thick,mark=*,mark options=solid] table {avg_overlap_sm.txt}; 
    
\end{axis}
\end{tikzpicture}

\begin{tikzpicture}
\begin{customlegend}[legend entries={Correlation-based tracker\cite{danelljan_accurate_2014}\,\,,STAPLE \cite{bertinetto_2016_staple}\,\,,Our approach},
        legend columns=3]
    \addlegendimage{yellow,fill=black!100!yellow,sharp plot}
    \addlegendimage{red,fill=black!100!red,sharp plot}
    \addlegendimage{green,fill=black!100!green,sharp plot}
    \end{customlegend}
\end{tikzpicture}
\hfill
\caption{The average bounding box overlap for rigid sequences within the VOT2016 and OTB-2015  datasets. We compare our overlap scores to STAPLE \cite{bertinetto_2016_staple} and a scale adaptive correlation tracker based on \cite{danelljan_accurate_2014}.}
\label{fig:scenesoverlap}
\end{figure}

To get a fair comparison, we compare our method to the state of the art STAPLE tracker \cite{bertinetto_2016_staple}, which was the best real-time tracker at the VOT-2016 challenge \cite{vot_2016}, and a very fast correlation filter tracker with scale adaption, based on \cite{danelljan_accurate_2014}. We evaluate the average bounding box overlap \cite{wu_otb_2015} on a selection of rigid objects from both the OTB-2015 and VOT-2016 datasets in \mbox{Fig. \ref{fig:scenesoverlap}}. Please note that the ground truth data is mostly only labeled as axis aligned bounding boxes which puts a heavy bias on the obtained overlaps.
For example, for the sequence {\tt Vase} we visually clearly outperform both approaches, as is seen nicely in \mbox{Fig. \ref{fig:examplesequence}}. Nevertheless, the bounding box abstraction lets the overlap drop very low. 

To measure the robustness, we do not use the VOT-2016 measures, but rather evaluate if the tracker is successfully tracking the target in the last few frames (bounding box overlap $>50\%$). Here STAPLE (22/26) and our approach (21/26) perform equally well, while the correlation based approach drops off (13/26).
In the following we will discuss individual sequences in more detail.

\begin{figure} [t]
\centering
\includegraphics[width=0.15\textwidth]{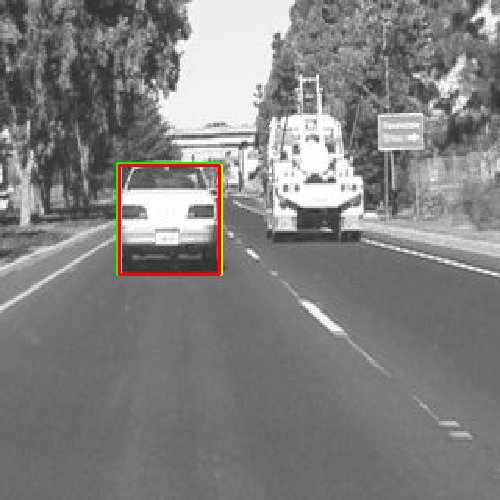}
\includegraphics[width=0.15\textwidth]{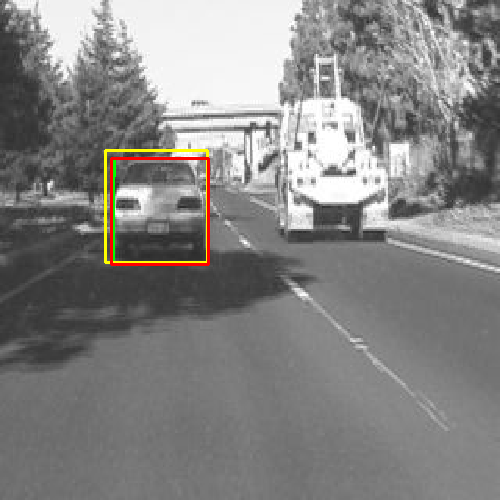}
\includegraphics[width=0.15\textwidth]{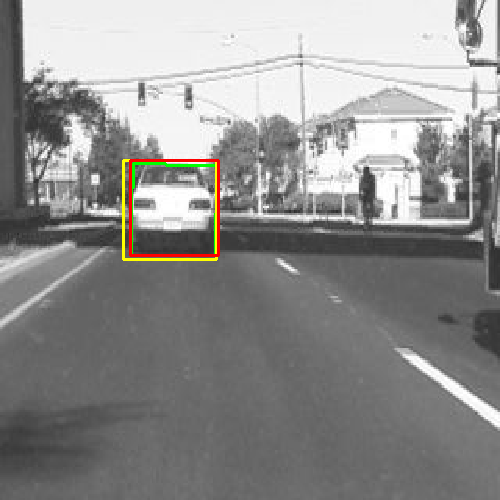}
\includegraphics[width=0.15\textwidth]{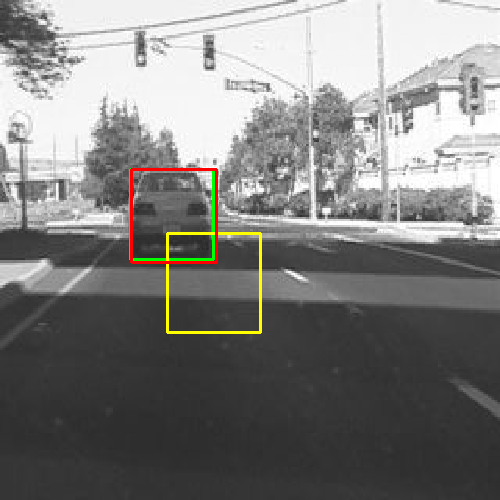}
\includegraphics[width=0.15\textwidth]{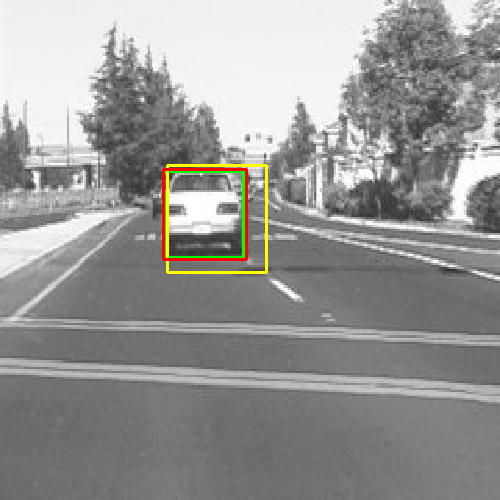}
\includegraphics[width=0.15\textwidth]{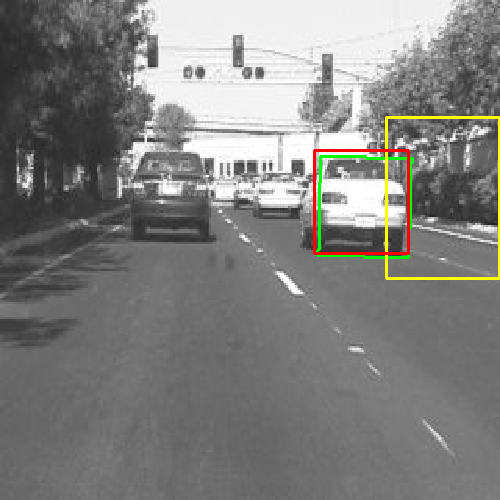}
\includegraphics[width=0.15\textwidth]{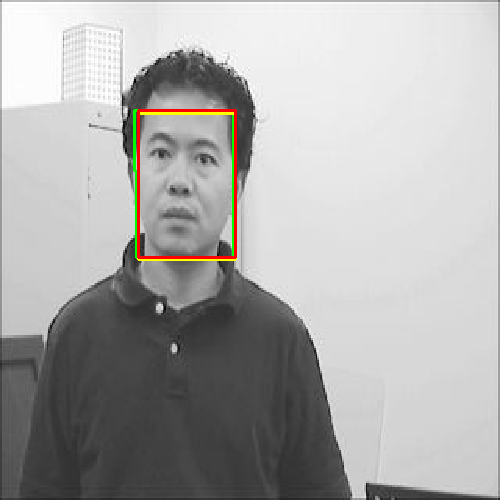}
\includegraphics[width=0.15\textwidth]{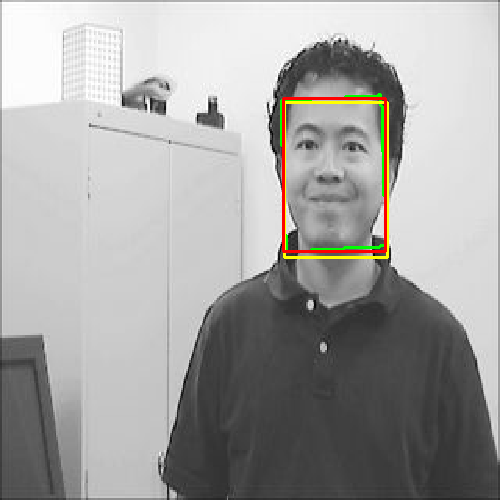}
\includegraphics[width=0.15\textwidth]{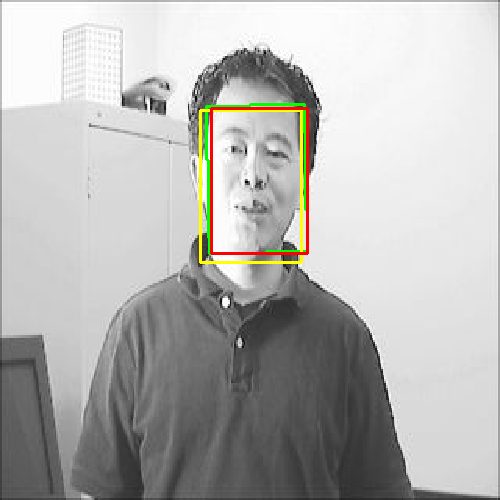}
\includegraphics[width=0.15\textwidth]{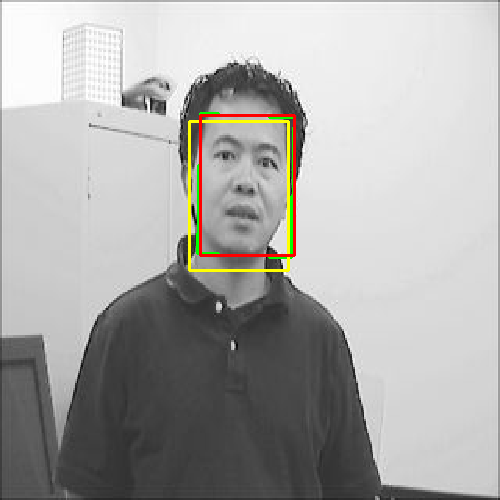}
\includegraphics[width=0.15\textwidth]{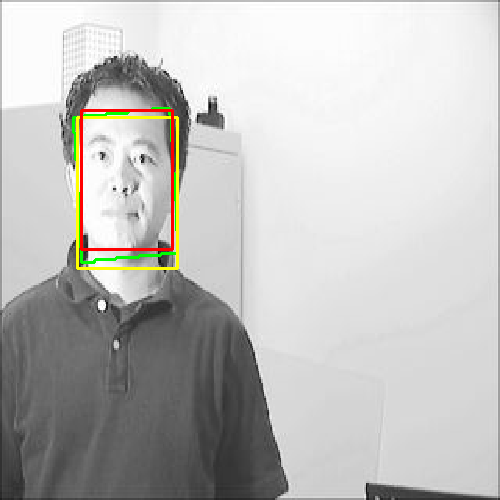}
\includegraphics[width=0.15\textwidth]{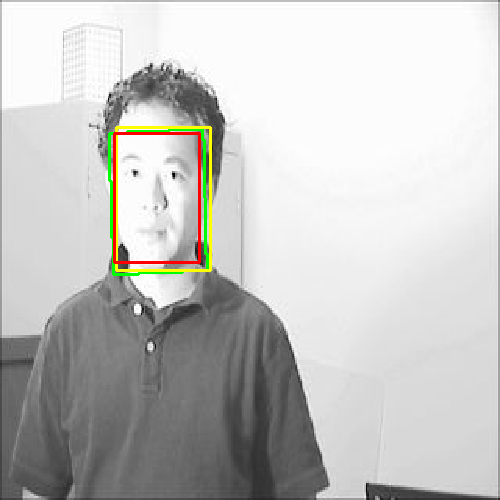}
\begin{tikzpicture}
\begin{customlegend}[legend entries={Correlation-based tracker\cite{danelljan_accurate_2014}\,\,,STAPLE\cite{bertinetto_2016_staple}\,\,,Our approach},
        legend columns=3]
    \addlegendimage{yellow,fill=black!100!yellow,sharp plot}
    \addlegendimage{red,fill=black!100!red,sharp plot}
    \addlegendimage{green,fill=black!100!green,sharp plot}
    \end{customlegend}
\end{tikzpicture}
\hfill
\caption{Our tracking scheme is inherently robust to illumination changes because of the similarity measure we use in the localization step \eqref{eq:similarity}. The method performs comparable to STAPLE and outperforms the scale adaptive correlation tracker, which fails in the first sequence, in terms of accuracy.}
\label{fig:exampleillumintaion}
\end{figure}

The choice of the similarity measure in \eqref{eq:similarity} makes the tracker inherently robust to nonlinear illumination changes. This is visualized for two sequences in \mbox{Fig. \ref{fig:exampleillumintaion}.}. The tracker localization quality is unaffected by the car driving under the bridge or the light being turned on and off in the second sequence. However, very strong changes of the illumination can make it difficult to segment the edge orientation of the target image robustly and lead to tracking failure, which is what happens in the sequence {\tt Fish}.

\begin{figure}[b]
\centering
\includegraphics[width=0.15\textwidth]{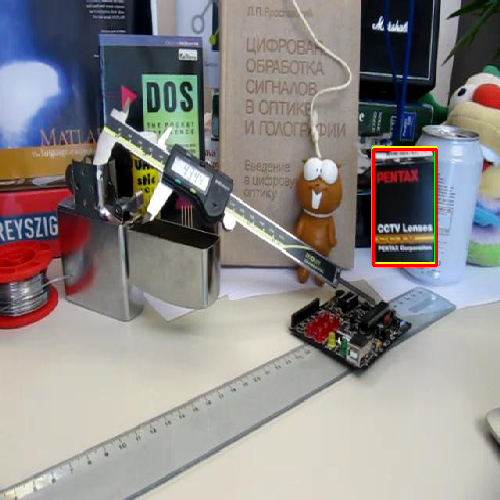}
\includegraphics[width=0.15\textwidth]{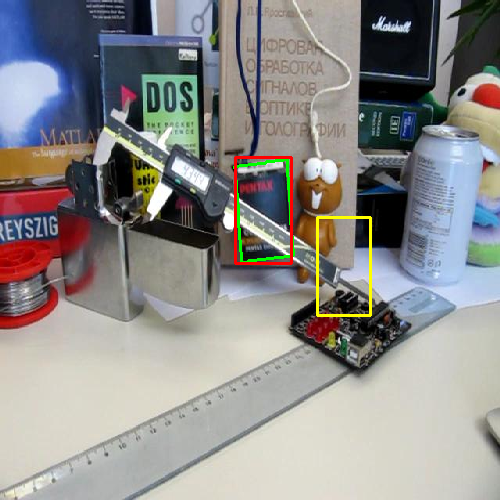}
\includegraphics[width=0.15\textwidth]{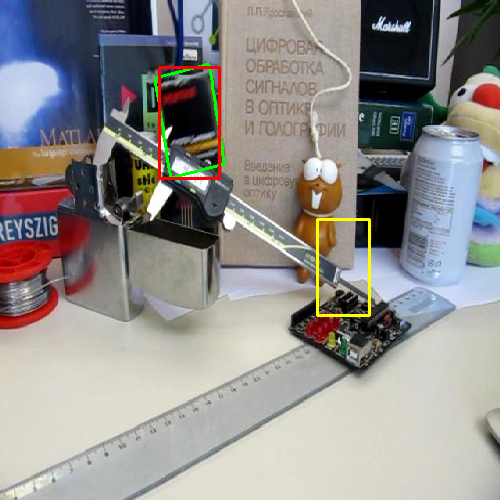}
\includegraphics[width=0.15\textwidth]{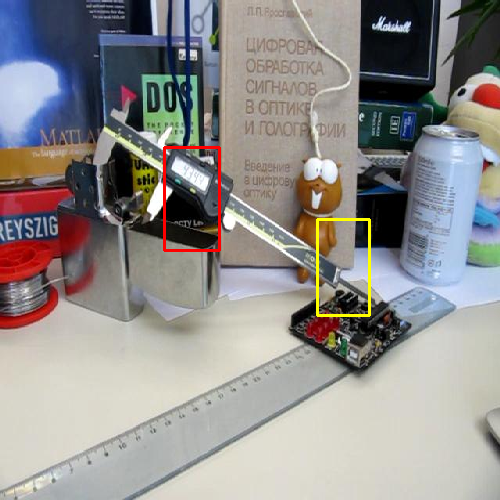}
\includegraphics[width=0.15\textwidth]{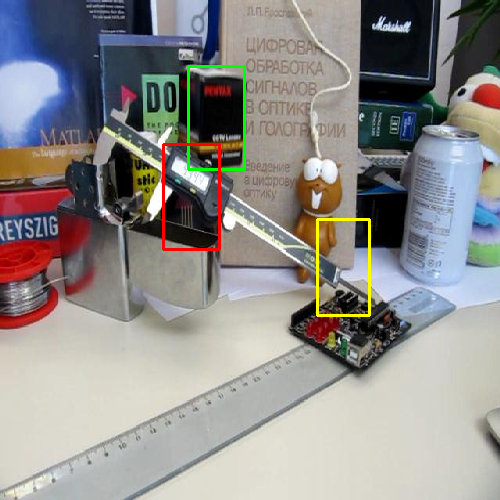}
\includegraphics[width=0.15\textwidth]{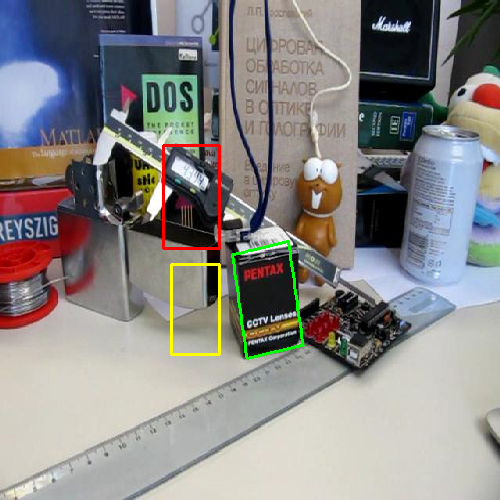}
\begin{tikzpicture}
\begin{customlegend}[legend entries={Correlation-based tracker\cite{danelljan_accurate_2014}\,\,,STAPLE\cite{bertinetto_2016_staple}\,\,,Our approach},
        legend columns=3]
    \addlegendimage{yellow,fill=black!100!yellow,sharp plot}
    \addlegendimage{red,fill=black!100!red,sharp plot}
    \addlegendimage{green,fill=black!100!green,sharp plot}
    \end{customlegend}
\end{tikzpicture}
\hfill
\caption{A further advantage of our approach is the possibility of self-diagnosing when the object is lost. The score is a reliable indicator of how much of the model is visible. In the sequence {\tt Box} from OTB-2015, all three trackers fail when the object is strongly occluded. Nevertheless, our approach recovers when the occluded object reappears.}
\label{fig:exampleredetection}
\end{figure}

The fact that our approach is essentially a local object detector with a meaningful score allows our framework to detect tracking failure reliably and recover from complete object occlusion. In \mbox{Fig. \ref{fig:exampleredetection},} the object is completely occluded in the middle of the sequence and hence the STAPLE and correlation-based tracker start adapting their filters to the foreground. Neither of the approaches is able to recover from the complete occlusion. Our approach, on the other hand, detects the tracking failure and is able to re-detect the object when it reemerges. 

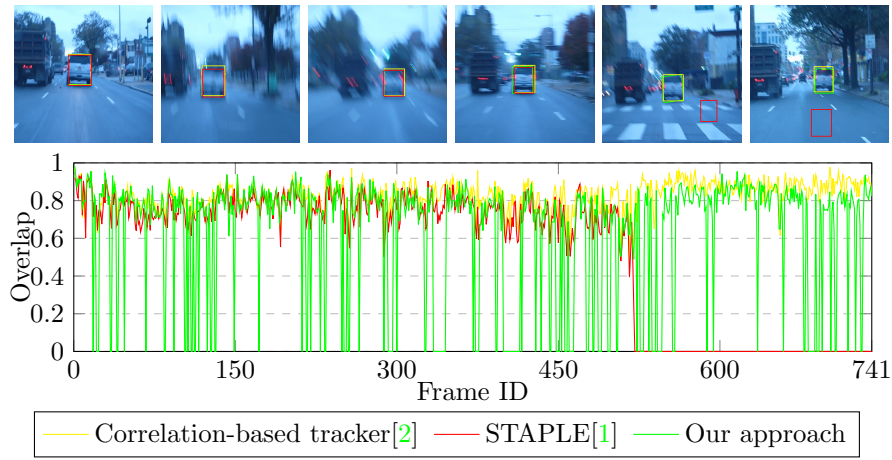
\begin{figure}
\centering
\includegraphics[width=0.15\textwidth]{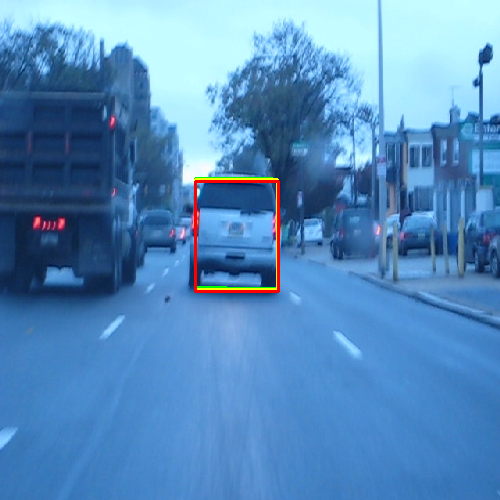}
\includegraphics[width=0.15\textwidth]{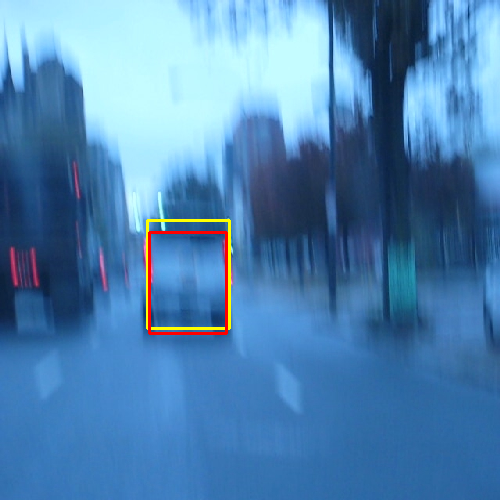}
\includegraphics[width=0.15\textwidth]{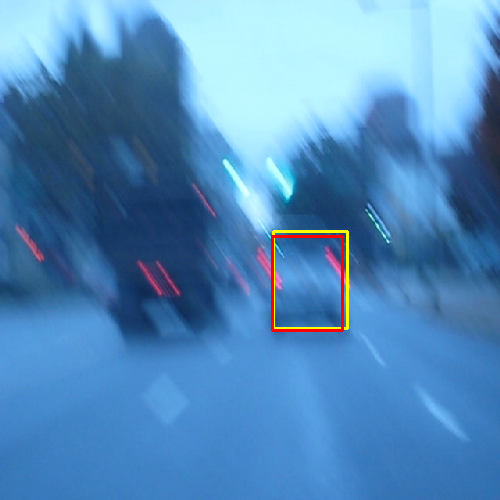}
\includegraphics[width=0.15\textwidth]{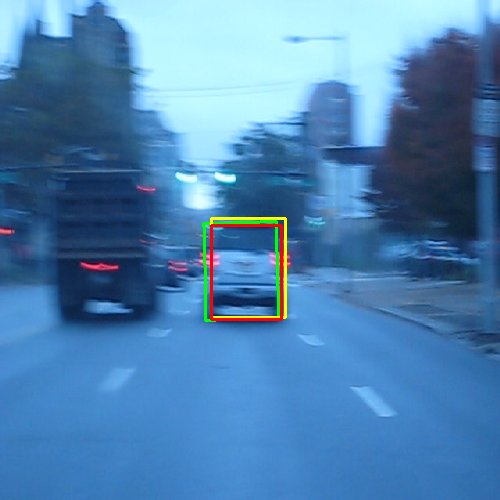}
\includegraphics[width=0.15\textwidth]{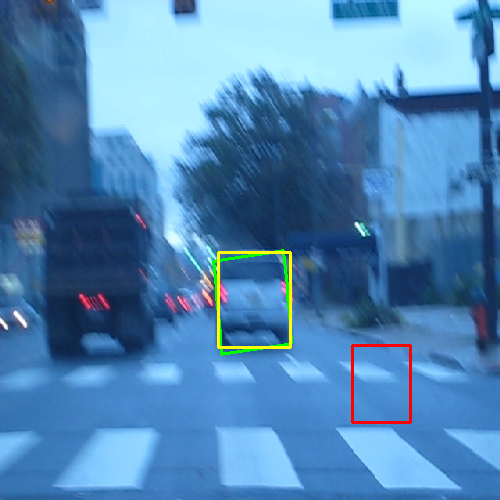}
\includegraphics[width=0.15\textwidth]{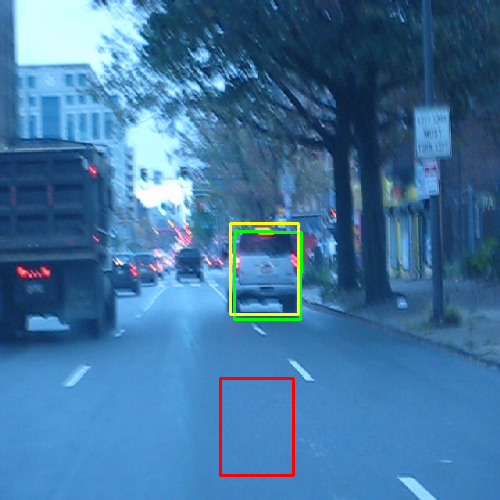}
\begin{tikzpicture}
\begin{axis}[
	scale only axis=true,
	width=10.5cm, height=2.5cm,
	x label style={at={(axis description cs:0.5,0.1)},anchor=north},
	y label style={at={(axis description cs:0.08,.5)},anchor=south},
    xlabel={Frame ID},
    ylabel={Overlap},
    xmin=0, xmax=741,
    ymin=0, ymax=1.00,
    xtick={0,150,300,450,600,741},
    ytick={0,.20,.40,.60,.80,1.00},
    legend pos=north west,
    ymajorgrids=true,
    grid style=dashed,
]
 
\addplot [color=yellow] table {overlap_kcf.dat};
\addplot [color=red] table {overlap_staple.dat};
\addplot [color=green] table {overlap_sm.dat}; 
    
\end{axis}
\end{tikzpicture}

\begin{tikzpicture}
\begin{customlegend}[legend entries={Correlation-based tracker\cite{danelljan_accurate_2014}\,\,,STAPLE\cite{bertinetto_2016_staple}\,\,,Our approach},
        legend columns=3]
    \addlegendimage{yellow,fill=black!100!yellow,sharp plot}
    \addlegendimage{red,fill=black!100!red,sharp plot}
    \addlegendimage{green,fill=black!100!green,sharp plot}
    \end{customlegend}
\end{tikzpicture}
\hfill
\caption{Edge-based tracking has difficulties when the image blur becomes too large. Neighboring edges may merge into each other or disappear completely. This becomes evident in the sequences {\tt BlurCar1} sequence from OTB-2015 \cite{wu_otb_2015}. Our tracker loses the object when the camera motion is too large. Nevertheless, in both sequences the tracker is always able to recover and finishes the sequence with a very good localization. The STAPLE tracker loses the track at frame 543 and does not recover.}
\label{fig:exampleblur}
\end{figure}

The fact that our approach searches for the best similarity transformation between the frames leads to the fact that our approach is weak in sequences where the object has strong local deformations. Furthermore, strong camera motion and image blur can make it fail. In the sequence in \mbox{Fig. \ref{fig:exampleblur}}, the object is lost whenever the camera motion is too strong. Fortunately, the object is always re-detected when the camera motion stops and the edges become clear enough. In the respective frame, the STAPLE tracker fails near the end when the camera motion is extremely high.

\section{Conclusion}
In this paper, we have proposed an efficient object tracker that is able to determine the position, scale \emph{and} rotation of a rigid objects in various different sequences with high accuracy. We validated our framework on a rigid subset of the VOT-2016 \cite{vot_2016} and OTB-2015 \cite{wu_otb_2015} datasets and were able to perform on par with real-time state-of-the art approaches in terms of robustness. As opposed to the existing schemes, our approach is more accurate in terms of localization. On the one hand this is due to the subpixel-precise refinement of the object pose and, on the other hand, due to estimating the object rotation.

Unfortunately, the label data of the existing benchmarks is restricted to axis-aligned and oriented bounding boxes, which makes it difficult to quantize the localization gains in the established performance measures. A subpixel-precise tracking dataset and evaluation framework that is not restricted to bounding boxes would be very helpful for future evaluations.

\bibliographystyle{plain}
\bibliography{shapemodel}

\begin{thebibliography}{10}

\bibitem{bertinetto_2016_staple}
Luca Bertinetto, Jack Valmadre, Stuart Golodetz, Ondrej Miksik, and Philip
  H.~S. Torr.
\newblock Staple: Complementary learners for real-time tracking.
\newblock In {\em {IEEE} {CVPR}}, pages 1401--1409, 2016.

\bibitem{danelljan_accurate_2014}
Martin Danelljan, Gustav H{\"{a}}ger, Fahad~Shahbaz Khan, and Michael Felsberg.
\newblock Accurate scale estimation for robust visual tracking.
\newblock In {\em {BMVC}}, 2014.

\bibitem{danelljan_beyond_2016}
Martin Danelljan, Andreas Robinson, Fahad~Shahbaz Khan, and Michael Felsberg.
\newblock Beyond correlation filters: Learning continuous convolution operators
  for visual tracking.
\newblock In {\em ECCV}, pages 472--488, 2016.

\bibitem{held_2016_learning}
David Held, Sebastian Thrun, and Silvio Savarese.
\newblock Learning to track at 100 {FPS} with deep regression networks.
\newblock In {\em {ECCV}}, pages 749--765, 2016.

\bibitem{henriques_exploiting_2012}
Jo{\~{a}}o~F. Henriques, Rui Caseiro, Pedro Martins, and Jorge Batista.
\newblock Exploiting the circulant structure of tracking-by-detection with
  kernels.
\newblock In {\em {ECCV}}, pages 702--715, 2012.

\bibitem{henriques_high_speed_2015}
Jo{\~{a}}o~F. Henriques, Rui Caseiro, Pedro Martins, and Jorge Batista.
\newblock High-speed tracking with kernelized correlation filters.
\newblock {\em {IEEE} Transactions on Pattern Analysis and Machine
  Intelligence}, 37(3):583--596, 2015.

\bibitem{vot_2016}
Matej Kristan, Ales Leonardis, Jiri Matas, Michael Felsberg, Roman~P.
  Pflugfelder, Luka Cehovin, Tom{\'{a}}s Voj{\'{\i}}r, and Gustav H{\"{a}}ger.
\newblock The visual object tracking {VOT2016} challenge results.
\newblock In {\em {ECCV} Workshops}, pages 777--823, 2016.

\bibitem{Kristan_2016_novel}
Matej Kristan, Jiri Matas, Ales Leonardis, Tom{\'{a}}s Voj{\'{\i}}r, Roman~P.
  Pflugfelder, Gustavo Fern{\'{a}}ndez, Georg Nebehay, Fatih Porikli, and Luka
  Cehovin.
\newblock A novel performance evaluation methodology for single-target
  trackers.
\newblock {\em {IEEE} Transactions on Pattern Analysis and Machine
  Intelligence}, 38(11):2137--2155, 2016.

\bibitem{lepetit_keypoint_2006}
Vincent Lepetit and Pascal Fua.
\newblock Keypoint recognition using randomized trees.
\newblock {\em {IEEE} Transactions on Pattern Analysis and Machine
  Intelligence}, 28(9):1465--1479, 2006.

\bibitem{mot_2016}
Anton Milan, Laura Leal{-}Taix{\'{e}}, Ian~D. Reid, Stefan Roth, and Konrad
  Schindler.
\newblock {MOT16:} {A} benchmark for multi-object tracking.
\newblock {\em CoRR}, abs/1603.00831, 2016.

\bibitem{nam_2016_learning}
Hyeonseob Nam and Bohyung Han.
\newblock Learning multi-domain convolutional neural networks for visual
  tracking.
\newblock In {\em {IEEE }CVPR}, pages 4293--4302, 2016.

\bibitem{steger_unbiased_1998}
Carsten Steger.
\newblock An unbiased detector of curvilinear structures.
\newblock {\em {IEEE} Transactions on Pattern Analysis and Machine
  Intelligence}, 20(2):113--125, 1998.

\bibitem{steger_similarity_2001}
Carsten Steger.
\newblock Similarity measures for occlusion, clutter, and illumination
  invariant object recognition.
\newblock In {\em Pattern Recognition, 23rd DAGM-Symposium, Proceedings}, pages
  148--154, 2001.

\bibitem{ulrich2011system}
M.~Ulrich and C.~Steger.
\newblock System and methods for automatic parameter determination in machine
  vision, May~31 2011.
\newblock US Patent 7,953,290.

\bibitem{ulrich_2002_performance}
Markus Ulrich and Carsten Steger.
\newblock Performance evaluation of 2d object recognition techniques.
\newblock Technical Report PF--2002--01, Lehrstuhl f{\"u}r Photogrammetrie und
  Fernerkundung, Technische Universit{\"a}t M{\"u}nchen, 2002.

\bibitem{wang_2015_visual}
Lijun Wang, Wanli Ouyang, Xiaogang Wang, and Huchuan Lu.
\newblock Visual tracking with fully convolutional networks.
\newblock In {\em {IEEE} {ICCV}}, pages 3119--3127, 2015.

\bibitem{wu_otb_2015}
Yi~Wu, Jongwoo Lim, and Ming{-}Hsuan Yang.
\newblock Object tracking benchmark.
\newblock {\em {IEEE} Transactions on Pattern Analysis and Machine
  Intelligence}, 37(9):1834--1848, 2015.

\bibitem{zhang_fast_2014}
Kaihua Zhang, Lei Zhang, Qingshan Liu, David Zhang, and Ming{-}Hsuan Yang.
\newblock Fast visual tracking via dense spatio-temporal context learning.
\newblock In {\em {ECCV}}, pages 127--141, 2014.

\end{thebibliography}

\end{document}